\documentclass{article}

\usepackage[preprint]{neurips_2023}

\usepackage[utf8]{inputenc} 
\usepackage[T1]{fontenc}    
\usepackage{hyperref}       
\usepackage{url}            
\usepackage{booktabs}       
\usepackage{amsfonts}       
\usepackage{nicefrac}       
\usepackage{microtype}      
\usepackage{xcolor}         

\usepackage{graphicx}
\usepackage{threeparttable}
\usepackage{makecell}
\usepackage{bm}
\usepackage{multirow}
\usepackage{enumitem}

\usepackage{caption}
\usepackage{subcaption}
\usepackage{CJKutf8}

\title{Can LLMs like GPT-4 outperform traditional AI tools in dementia diagnosis? Maybe, but not today}

\author{Zhuo Wang\textsuperscript{\rm 1}, Rongzhen Li\textsuperscript{\rm 2}, Bowen Dong\textsuperscript{\rm 1}, Jie Wang\textsuperscript{\rm 3}, Xiuxing Li\textsuperscript{\rm 4,5}, Ning Liu\textsuperscript{\rm 7}, \\\textbf{Chenhui Mao}\textsuperscript{\rm 2}, \textbf{Wei Zhang}\textsuperscript{\rm 6}, \textbf{Liling Dong}\textsuperscript{\rm 2}, \textbf{Jing Gao}\textsuperscript{\rm 2}\footnotemark[1], \textbf{Jianyong Wang}\textsuperscript{\rm 1}\thanks{Corresponding authors.}\footnotemark[1]\\ 
\textsuperscript{\rm 1}Department of Computer Science and Technology, Tsinghua University\\
\textsuperscript{\rm 2}Department of Neurology, State Key Laboratory of Complex Severe and Rare \\Diseases, Peking Union Medical College Hospital, Chinese Academy of \\Medical Science and Peking Union Medical College\\
\textsuperscript{\rm 3}The First Affiliated Hospital of Nanchang University\\
\textsuperscript{\rm 4}Key Laboratory of Intelligent Information Processing Institute of \\Computing Technology, Chinese Academy of Sciences (ICT/CAS)\\
\textsuperscript{\rm 5}University of Chinese Academy of Sciences\\
\textsuperscript{\rm 6}School of Computer Science and Technology, East China Normal University\\
\textsuperscript{\rm 7}School of Software, Shandong University\\
\{wang-z18, dbw22\}@mails.tsinghua.edu.cn, \\\{lirongzhen18, wangjie\_smu, sophie\_d\}@163.com, \\lixiuxing@ict.ac.cn, \{victorliucs, zhangwei.thu2011\}@gmail.com,\\ maochenhui@pumch.cn, gj107@163.com, jianyong@tsinghua.edu.cn 
}

\begin{document}

\maketitle

\begin{abstract}
Recent investigations show that large language models (LLMs), specifically GPT-4, not only have remarkable capabilities in common Natural Language Processing (NLP) tasks but also exhibit human-level performance on various professional and academic benchmarks.
However, whether GPT-4 can be directly used in practical applications and replace traditional artificial intelligence (AI) tools in specialized domains requires further experimental validation.
In this paper, we explore the potential of LLMs such as GPT-4 to outperform traditional AI tools in dementia diagnosis.
Comprehensive comparisons between GPT-4 and traditional AI tools are conducted to examine their diagnostic accuracy in a clinical setting. 
Experimental results on two real clinical datasets show that, although LLMs like GPT-4 demonstrate potential for future advancements in dementia diagnosis, they currently do not surpass the performance of traditional AI tools. 
The interpretability and faithfulness of GPT-4 are also evaluated by comparison with real doctors.
We discuss the limitations of GPT-4 in its current state and propose future research directions to enhance GPT-4 in dementia diagnosis.
\end{abstract}

\section{Introduction}
In recent years, Large Language Models (LLMs), powered by advanced deep learning techniques and massive cross-disciplinary corpora, have significantly impacted the field of Natural Language Processing (NLP) and achieved great success in a wide range of NLP tasks \citep{zhao2023survey,devlin2018bert,brown2020language,fan2023bibliometric}. 
As one of the most powerful LLMs, GPT-4 \citep{openai2023gpt4}, a transformer-based language model, has advanced the field of NLP even further. With its remarkable ability to comprehend and generate coherent and contextually relevant text, GPT-4 has become a powerful tool for various tasks, including machine translation, sentiment analysis, and question-answering systems \citep{bubeck2023sparks}. 
Recent investigations show that, besides the above common NLP tasks, GPT-4 also exhibit human-level performance on various professional and academic benchmarks \citep{bubeck2023sparks,openai2023gpt4,nori2023capabilities}.
For example, GPT-4 achieves a score that falls in the top 10\% of test takers on a simulated bar exam and exceeds the passing score on the United States Medical Licensing Examination (USMLE) by over 20 points without any specialized prompt crafting.

The impressive performance of GPT-4 on various professional and academic benchmarks has prompted practitioners to explore the potential of GPT-4 (or its predecessor, e.g., GPT-3.5) in practical applications in specialized domains \citep{vaghefi2023chatipcc,perlis2023application,vidgof2023large,park2023correct}, e.g., clinical medicine.
However, given the complexity and specificity of such practical applications, it remains uncertain how effective GPT-4 could be in these contexts. Therefore, further experimentation is warranted to verify the capacity and potential impact of GPT-4.
In this paper, we focus on the task of dementia diagnosis and seek to answer the following question: \textit{Can GPT-4 fulfil the requirements of dementia diagnosis and replace traditional AI tools in this task?}

Dementia, as one of the major causes of disability and dependency among the elderly population, has garnered increasing attention and concern \citep{whodementia,world2019risk,patterson2018world}. With no cure for dementia currently available, early intervention is the most effective approach.
However, early diagnosis and progression prediction remain challenging, with low accuracy, leading to most patients being diagnosed after having severe symptoms, i.e., in the later stage of dementia, when the best time for the interventions has already passed \citep{olazaran2010nonpharmacological,prince2018world}.

Traditional AI tools, such as supervised learning models, have been proposed to improve the performance of dementia early diagnosis and prediction \citep{wang2022random,wang2022learning,palmqvist2021prediction}. The advantage of these models is that they can extract implicit new knowledge from the data, which may not appear in the existing literature. However, these models also have their shortcomings. Firstly, they require the collection of training datasets, which can be time-consuming and labour-intensive, with the quality of the training data having a significant impact on the final performance. Secondly, most of the effective machine learning models, e.g., deep learning and ensemble models, are black-box models \citep{doshi2017towards,molnar2020interpretable,lipton2018mythos}. Since we can hardly understand their decision mechanism, potential biases and risks could hide in these models. It is also challenging for these black-box models to assist doctors in diagnosis. Lastly, these conventional AI tools lack the capacity to leverage the knowledge contained in other data sources or corpora like LLMs.

Unlike traditional machine learning methods that require a specific training dataset, LLMs like GPT-4 leverage their extensive knowledge acquired from massive cross-disciplinary corpora, which may enable them to obtain promising results even in zero-shot or few-shot settings \citep{nori2023capabilities}. 
This advantage eliminates the need to collect specialized training sets, significantly reducing the time and resources required for diagnostic model development.
Furthermore, GPT-4 has the ability to provide interpretable explanations for its decisions, allowing doctors to gain insights into the underlying reasoning process \citep{openai2023gpt4,lee2023benefits}.
Despite these promising aspects, there are still several questions to be answered when utilizing GPT-4 for dementia diagnosis. 
The first question is how to design a simple but effective prompt template for dementia diagnosis. In addition, without fine-tuning, the performance of zero-shot and few-shot learning for dementia diagnosis is unknown.

In this paper, we aim to answer the above questions and explore the potential of GPT-4 in dementia diagnosis. We summarize the key contributions as follows:
\begin{itemize}[leftmargin=*,itemsep=0pt,parsep=0.0em,topsep=0.0em,partopsep=0.0em]
\item We design simple but effective prompt templates of GPT-4 for dementia diagnosis.
\item We investigate the capabilities of GPT-4 on dementia diagnosis by comprehensively comparing GPT-4 with traditional AI tools and doctors on two real clinical datasets.
\item We identify the limitations and challenges faced by GPT-4 in the context of dementia diagnosis and discuss possible directions for future work.
\end{itemize}

\section{Materials and Methods}
\subsection{Data origin and acquisition}
\begin{table*}[h!]
\begin{threeparttable}
\renewcommand{\arraystretch}{2.5}
\caption{Demographic characteristics of the ADNI and PUMCH datasets.}
\label{tab:demographics}
  \begin{tabular}{ccccccccc}
  \hline
 & 		 \multicolumn{3}{c}{\textbf{ADNI}}	 & 		 & \multicolumn{4}{c}{\textbf{PUMCH}} 	 \\ \cline{2-4} \cline{6-9}
	 & 	\makecell{\textbf{MCI-NC} \\ \textbf{n=253}}	 & 	\makecell{\textbf{MCI-C} \\ \textbf{n=353}} & \textbf{P value}  && \makecell{\textbf{CN} \\ \textbf{n=67}}	&	\makecell{\textbf{MCI} \\ \textbf{n=174}}	& \makecell{\textbf{Dementia} \\ \textbf{n=134}} & \textbf{P value}\\ \hline	

\makecell{\textbf{Age}\\\textbf{ (years)}}	&	\makecell{70.83\\(7.26)}	&	\makecell{73.89\\(7.11)} 	&	 $<$0.001 	&		&	 \makecell{63.24\\(12.00)}	&	\makecell{64.16\\(11.61)}	&	\makecell{68.41\\(10.44)} 	&	 $<$0.001	\\
\makecell{\textbf{Gender}\\\textbf{ (female)}}	&	\makecell{102\\(40.3\%)}	&	\makecell{140\\(39.7\%)} 	&	0.937	&		&	 \makecell{43\\(64.2\%)}	&	\makecell{99\\(56.9\%)}	&	\makecell{72\\(53.7\%)} 	&	0.371	\\
\makecell{\textbf{Education}\\\textbf{ (years)}}	&	\makecell{16.11\\(2.78)}	&	\makecell{15.91\\(2.75)} 	&	0.384	&		&	 \makecell{13.88\\(3.34)}	&	\makecell{11.93\\(3.98)}	&	\makecell{11.96\\(3.92)} 	&	$<$0.002	\\
\makecell{\textbf{MMSE}}	&	\makecell{28.24\\(1.61)}	&	\makecell{27.07\\(1.76)} 	&	$<$0.001 	&		&	 \makecell{28.70\\(1.17)}	&	\makecell{27.95\\(1.22)}	&	\makecell{27.15\\(1.17)} 	&	 $<$0.001	\\
\makecell{\textbf{MoCA}}	&	\makecell{24.35\\(2.76)}	&	\makecell{21.73\\(2.79)} 	&	$<$0.001 	&		&	 \makecell{27.18\\(1.65)}	&	\makecell{24.64\\(2.77)}	&	\makecell{22.54\\(2.82)} 	&	 $<$0.001	\\
    \hline
 \end{tabular}
\begin{tablenotes}[flushleft]
\item []
Data are shown as mean (s.d.) or n (\%). Abbreviations: MCI = mild cognitive impairment; MCI-C = MCI converter; MCI-NC = MCI non-converter; CN = Cognitively Normal;  MMSE = Mini-Mental State Examination; MoCA = Montreal Cognitive Assessment.
\end{tablenotes}
\end{threeparttable}
\end{table*}

This study utilizes two distinct datasets. The first dataset is sourced from the Alzheimer's Disease Neuroimaging Initiative (ADNI) database (adni.loni.usc.edu), which includes ADNI 1, 2/GO, and 3 \citep{petersen2010alzheimer}.
The ADNI is a longitudinal multicenter study designed to develop clinical, imaging, genetic, and biochemical biomarkers for the early detection and tracking of Alzheimer’s disease (AD).
The primary objective of using the ADNI dataset is to distinguish between patients with mild cognitive impairment (MCI) who develop AD, i.e., MCI converters (MCI-C), and those with MCI who do not develop AD, i.e., MCI non-converters (MCI-NC).
Subjects are included consecutively. 
After pre-processing the original data and eliminating invalid records, 606 participants remain, with 253 (41.7\%) MCI-NC and 353 (58.3\%) MCI-C.
It is confirmed that all MCI-NC patients do not progress to AD after at least 48 months of follow-up.
Each subject has 51 features, including the demographic information, the results of selected cognitive tests (e.g., MMSE score), and other biomarkers (e.g., APOE4, AV45, and pTau).

Considering that the training dataset of GPT-4 may contain information from the publicly available ADNI dataset, potentially leading to information leakage issues, the second dataset used in this study is a private dataset.
The second dataset was collected by the Peking Union Medical College Hospital (PUMCH) from May 2009 to April 2021 \citep{wang2022random}.
Inclusion criteria require subjects to have a normal MMSE score ($\ge 26$) and the capability to complete all required neuropsychological assessments. Subjects are included consecutively. 
Diagnoses are determined using clinical history, neuropsychological tests, laboratory tests, and head CT or MRI scans.
A total of 375 subjects are included, among which 67 (17.9\%) subjects are diagnosed with cognitively normal (CN), 174 (46.4\%) are diagnosed with MCI, and 134 (35.7\%) are diagnosed with dementia. CN and MCI are collectively referred to as non-dementia.
We use PUMCH-B to represent the binary classification tasks (Non-Dementia vs. Dementia), and PUMCH-T to represent the ternary classification task (CN vs. MCI vs. Dementia).
The demographic information and the results of selected cognitive tests in each record are converted into 64 features after data preprocessing.

The demographic characteristics of the ADNI and PUMCH datasets are shown in Table \ref{tab:demographics}.

\begin{figure*}
    \centering
    \includegraphics[width=0.7\textwidth]{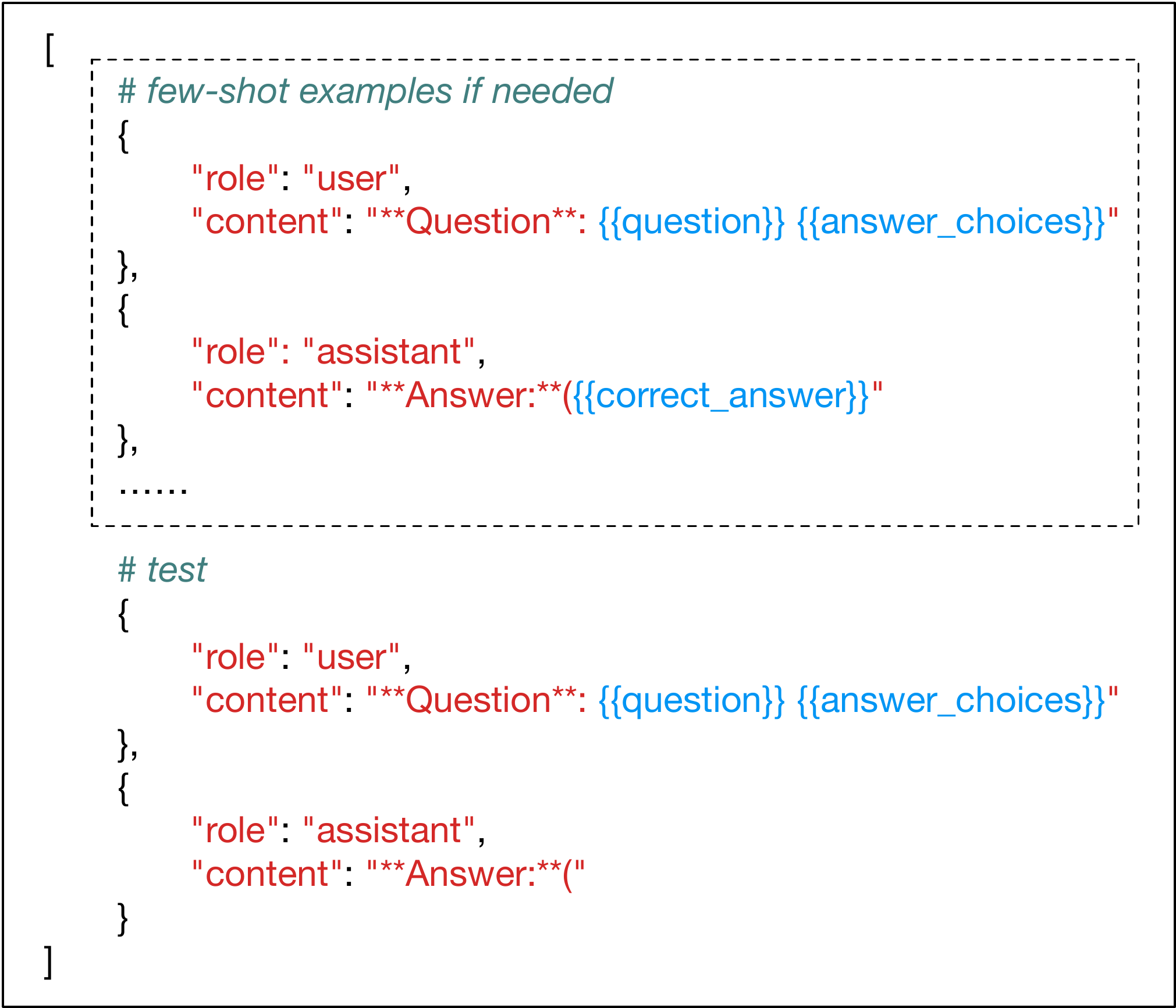}
    \caption{Template used to generate prompts for GPT-4 by converting the dementia diagnosis into a multiple choice question. Elements in double braces \{\{\}\} are replaced with question-specific values. The content in the dashed box is optional.}
    \label{fig:prompt_template}
\end{figure*}

\begin{figure}
     \centering
     \begin{subfigure}[b]{0.47\textwidth}
         \centering
         \includegraphics[width=\textwidth]{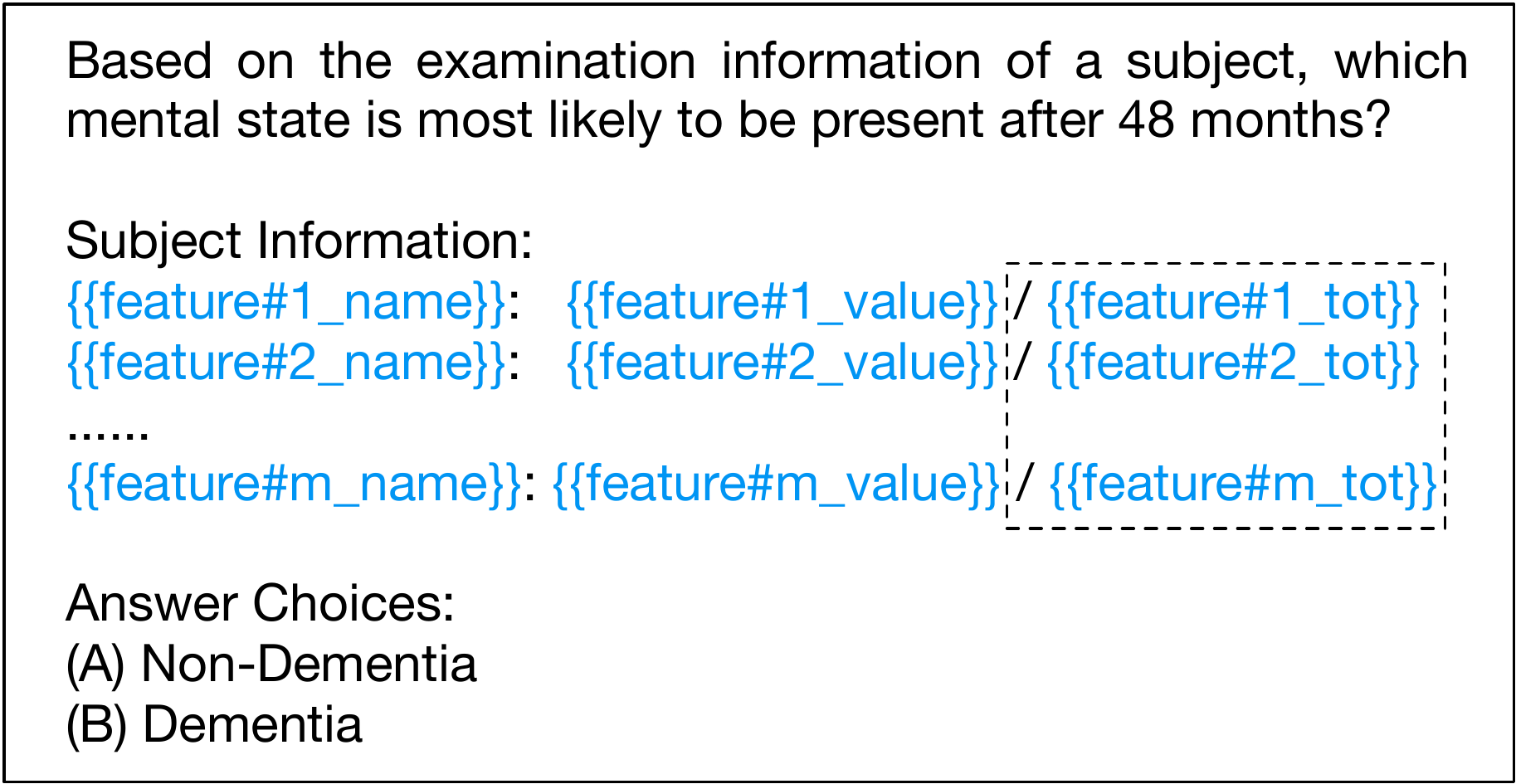}
         \caption{Template for ADNI}
         \label{fig:adni_prompt}
     \end{subfigure}
     \begin{subfigure}[b]{0.52\textwidth}
         \centering
         \includegraphics[width=\textwidth]{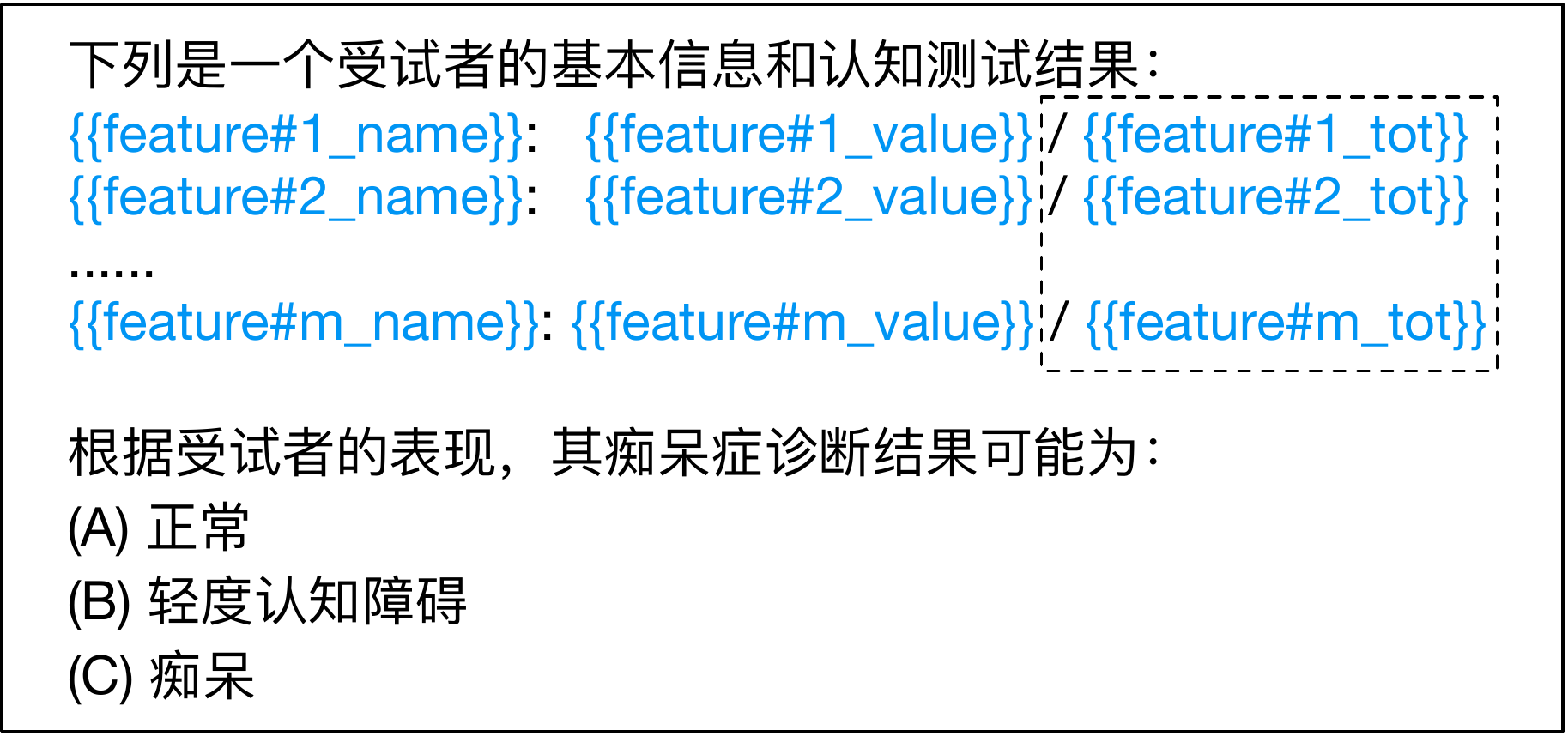}
         \caption{Template for PUMCH-T}
         \label{fig:pumch_prompt}
     \end{subfigure}
     \caption{The templates of questions and answer choices for different datasets. The content in the dashed box is optional.}
        \label{fig:both_prompt}
\end{figure}

\subsection{Prompting}
A prompt is an input query or context that guides the LLMs to generate relevant and meaningful responses. 
Prompts are essential for directing the LLM towards the desired information or output, leveraging the vast knowledge of LLM effectively.
To obtain accurate and contextually appropriate responses from LLMs by developing and optimizing prompts, prompt engineering is needed.

Considering that the demographic information, the results of selected cognitive tests, and other biomarkers are all provided for the diagnosis of dementia, we employ the prompt template shown in Figure \ref{fig:prompt_template}. 
This prompt template converts the dementia diagnosis (prediction) into a multiple-choice question format. Therefore, we can obtain the answer by creating a completion for the provided prompt using LLMs. 
The content in the dashed box is optional, and it enables us to seek better performance by few-shot learning.
For different datasets (tasks), we also design different templates of questions and answer choices. Figure \ref{fig:adni_prompt} and \ref{fig:pumch_prompt} are the templates (i.e., \{\{question\}\} \{\{answer\_choice\}\} in Figure \ref{fig:prompt_template}) for the ADNI and PUMCH-T, respectively. The templates for PUMCH-B and PUMCH-T are similar.
In the template, we list the feature name and the corresponding feature value one by one. 
Since one test may correspond to different standards, to prevent confusion, we not only have doctors standardize feature names but also provide information such as total scores.
Figure \ref{fig:pumch_case_1} and \ref{fig:pumch_case_2} show two examples of the PUMCH-T template.

Furthermore, when using OpenAI's API, we can also set some parameters to constrain the responses of GPT-4 and GPT-3.5. For instance, we can set the parameter \textit{max\_tokens} to 1 to make GPT-4 respond only options. Additionally, we can use the parameter \textit{logit\_bias} to modify the likelihood of specified tokens appearing in the completion, thus making GPT-4 respond with only options A, B, and C. We can also adjust the parameter \textit{temperature} to make GPT-4 more focused and deterministic.

It should be noted that each data set we used is actually a table, and each row represents one subject (instance) while each column represents one feature in those tables. Although GPT-4 supports the inputs in a table format, its capability of handling table data input is quite poor compared with handling the input using our prompt template, especially under the few-shot learning settings. Therefore, although a table format input is simpler and shorter than an input using our template, we still use the proposed template to help GPT-4 understand and obtain a better performance. Our experiments also verify that the proposed template can help GPT-4 remember the few-shot examples.




\subsection{Evaluation}
We adopt accuracy to evaluate the classification performance. For each dataset, we randomly select 90\% as the training set and use the remaining 10\% as the test set.
We split the dataset according to the roster ID. Therefore, no patient is included in both the training and test sets, and the risk of data leakage in supervised learning is avoided \citep{saravanan2018data}. 

\subsection{Model Comparison}
The performance of GPT-4 and its predecessor model, GPT-3.5, is compared with five representative supervised machine learning models, including interpretable models and complex models (black-box models) that are hard to interpret and understand. CART \citep{breiman2017classification} is a rule-based model that builds a decision tree. Logistic Regression (LR) \citep{kleinbaum2002logistic} is a linear model. Rule-based Representation Learner (RRL) \citep{wang2021scalable} uses neural networks to learn interpretable rules. These three models are considered interpretable models. 
Random Forest (RF) \citep{breiman2001random} and eXtreme Gradient Boosting (XGBoost) \citep{chen2016xgboost} are considered complex models since they are ensemble models consisting of hundreds of decision trees. RF and XGBoost are hard to interpret due to their complex inner structures.

\section{Results}
\subsection{Classification Performance}
We compare the classification accuracy of GPT-4 and GPT-3.5 with five representative machine learning models. 
The results are shown in Table \ref{tab:accuracy}.
We can observe that the supervised model RRL consistently outperforms all other models across all datasets, verifying its good capability in dementia diagnosis.
Although GPT-4 shows promising results, it still has a noticeable performance gap compared to RRL, particularly on the PUMCH-T dataset.
Although GPT-4 outperforms simple models like LR and DT in accuracy in some cases, it could not entirely replace them, possibly due to the limitations of zero-shot or few-shot learning.
Consequently, GPT-4 is far from replacing more effective models like RRL in dementia diagnosis tasks. 

We can also see that GPT-4 exhibits a significant improvement over GPT-3.5, particularly on the ADNI dataset. However, we cannot definitively rule out the possibility of information leakage in GPT-4 (i.e., the whole or part of the ADNI dataset is included in GPT-4). Considering the substantial improvements observed in the private datasets (i.e., PUMCH-B and PUMCH-T), GPT-4 is indeed more suitable and powerful for dementia diagnosis than GPT-3.5.
For both GPT-4 and GPT-3.5, few-shot learning settings could have better results compared to zero-shot learning settings in some cases, indicating the potential benefits of providing additional context.

\begin{table*}[h!]
\scalebox{0.93}{
\begin{threeparttable}
\renewcommand{\arraystretch}{1.75}
\caption{Accuracy of comparing models on the ADNI and PUMCH datasets.}
\label{tab:accuracy}
  \begin{tabular}{ccccccccccc}
  \toprule
	&	\multicolumn{2}{c}{GPT-4}			&	&	\multicolumn{2}{c}{GPT-3.5}			&	\multirow{2}{*}{LR}	&	\multirow{2}{*}{DT}	&	\multirow{2}{*}{RF}	&	\multirow{2}{*}{XGBoost}	&	\multirow{2}{*}{RRL}	\\\cline{2-3} \cline{5-6}
	&	0-shot	&	few-shot	&	&	0-shot	&	few-shot	&		&		&		&		&		\\\hline
ADNI	&	0.820	&	0.820	&	&	0.443	&	0.639	&	0.803	&	0.689	&	0.803	&	0.820	&	\textbf{0.852}	\\
PUMCH-B	&	0.737	&	0.737	&	&	0.632	&	0.658	&	0.684	&	0.737	&	0.711	&	0.737	&	\textbf{0.789}	\\
PUMCH-T	&	0.553	&	0.632	&	&	0.474	&	0.553	&	0.526	&	0.684	&	0.711	&	0.711	&	\textbf{0.763}	\\
    \bottomrule
 \end{tabular}
\begin{tablenotes}[flushleft]
\item []
\end{tablenotes}
\end{threeparttable}
}
\end{table*}


\subsection{Case Study}
We show how the diagnosis generated by GPT-4 looks like by case studies. 
By comparing GPT-4 with professional doctors, we can not only intuitively understand the differences between them, but also qualitatively evaluate the \textbf{interpretability} and \textbf{faithfulness} of GPT-4. 
In addition, due to the inability of doctors to perform dementia prediction tasks, we only compared them on the PUMCH dataset.

\begin{figure*}
    \centering
    \includegraphics[width=\textwidth]{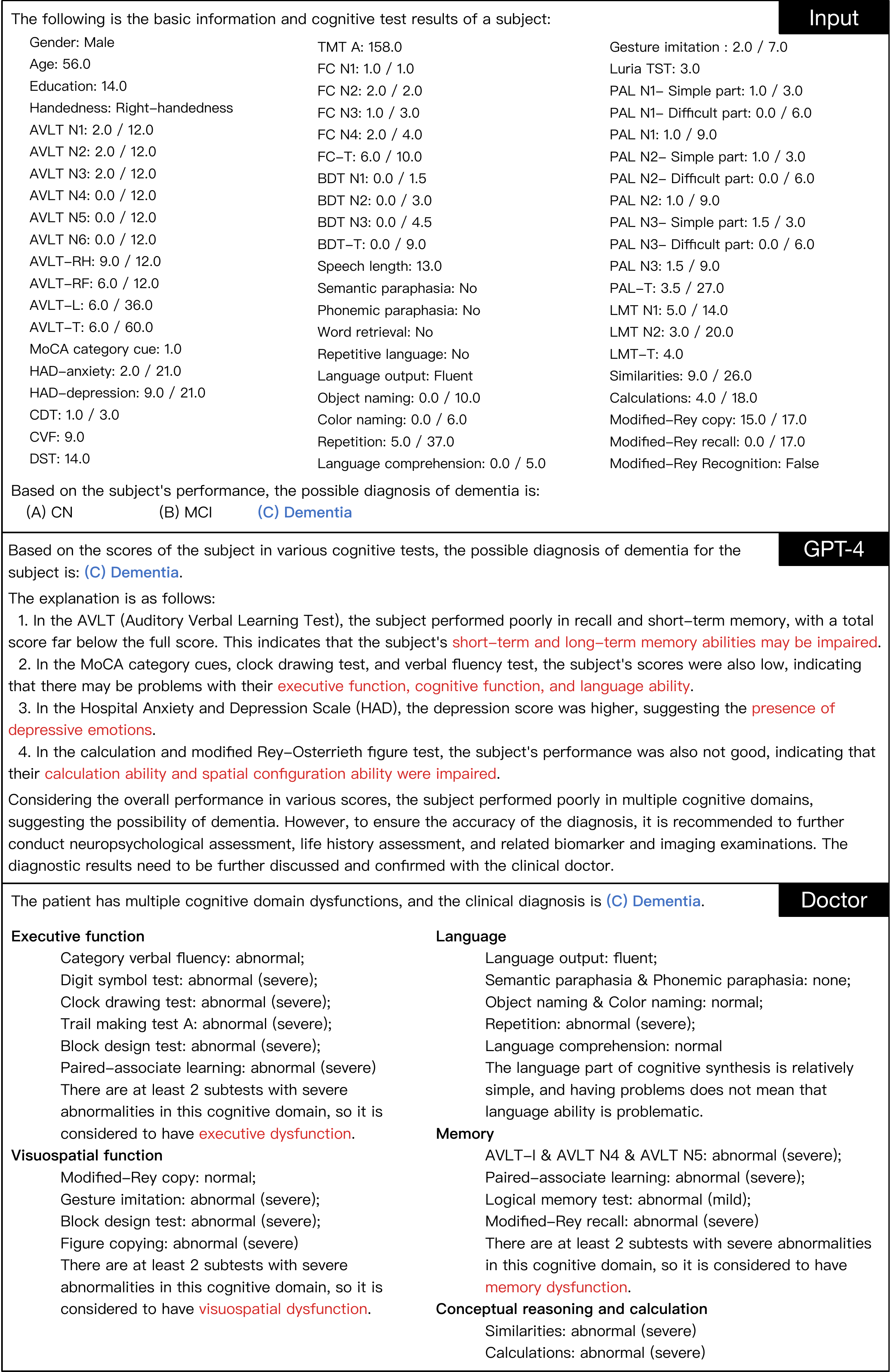}
    \caption{The first example of a comparison between GPT-4 and a doctor's diagnosis on the PUMCH dataset (English translation). The Chinese original text is shown in Figure \ref{fig:pumch_case_cn_1}, and a detailed description of each feature can be found in Table \ref{tab:feature-table}.}
    \label{fig:pumch_case_1}
\end{figure*}

Figure \ref{fig:pumch_case_1} shows the first example of a comparison between GPT-4 and a doctor's diagnosis. It is important to emphasize that PUMCH is a Chinese dataset, and Figure \ref{fig:pumch_case_1} shows the content translated from the Chinese original text. The Chinese original text is shown in Figure \ref{fig:pumch_case_cn_1} in the appendix. The first part of Figure \ref{fig:pumch_case_1} shows an example of input using the template shown in Figure \ref{fig:both_prompt}. The blue-highlighted option is the ground truth label for this example. This part provides a detailed display of the subject's basic information and cognitive test results (i.e., features). For ease of presentation, we only show abbreviations for feature names in the English translation version, while the detailed names are used in the Chinese original version. The detailed description of each feature can be found in Table \ref{tab:feature-table} in the appendix. The second part of the figure shows the diagnostic results of GPT-4. The blue-highlighted part is GPT-4's final conclusion, and the red-highlighted part is the cognitive function that GPT-4 believes may be related to dementia and have potential issues. To make GPT-4 explain its decision, we can add a sentence like "Please provide a detailed explanation" at the end of the input. The third part of the figure shows the doctor's diagnostic results. Similarly, the blue-highlighted part is the doctor's final conclusion, and the red-highlighted part is the cognitive function that the doctor believes has issues.

We can see that for the first example, both GPT-4 and the doctor diagnose the subject as having dementia, but the explanations for the diagnosis are different. First, we can see that both GPT-4 and the doctor explain their decisions using natural language. Since these sentences are easy to read and understand, the interpretability of GPT-4 is good. Second, we can observe that GPT-4 analyzes the input sequentially (as the explanation shows) and then summarizes the results, while the doctor analyzes the input according to the cognitive functions and then integrates the results. In comparison, the doctor's diagnostic approach is more in line with human understanding, and its readability and interpretability are better. In addition, both GPT-4 and the doctor point out that the subject has problems in executive function, visuospatial function, memory, and calculation. GPT-4 also emphasizes the presence of depressive emotions. This indicates the consistency between GPT-4 and the doctor is relatively high, indirectly verifying the good faithfulness of the explanation provided by GPT-4 in this case.

\begin{figure*}
    \centering
    \includegraphics[width=\textwidth]{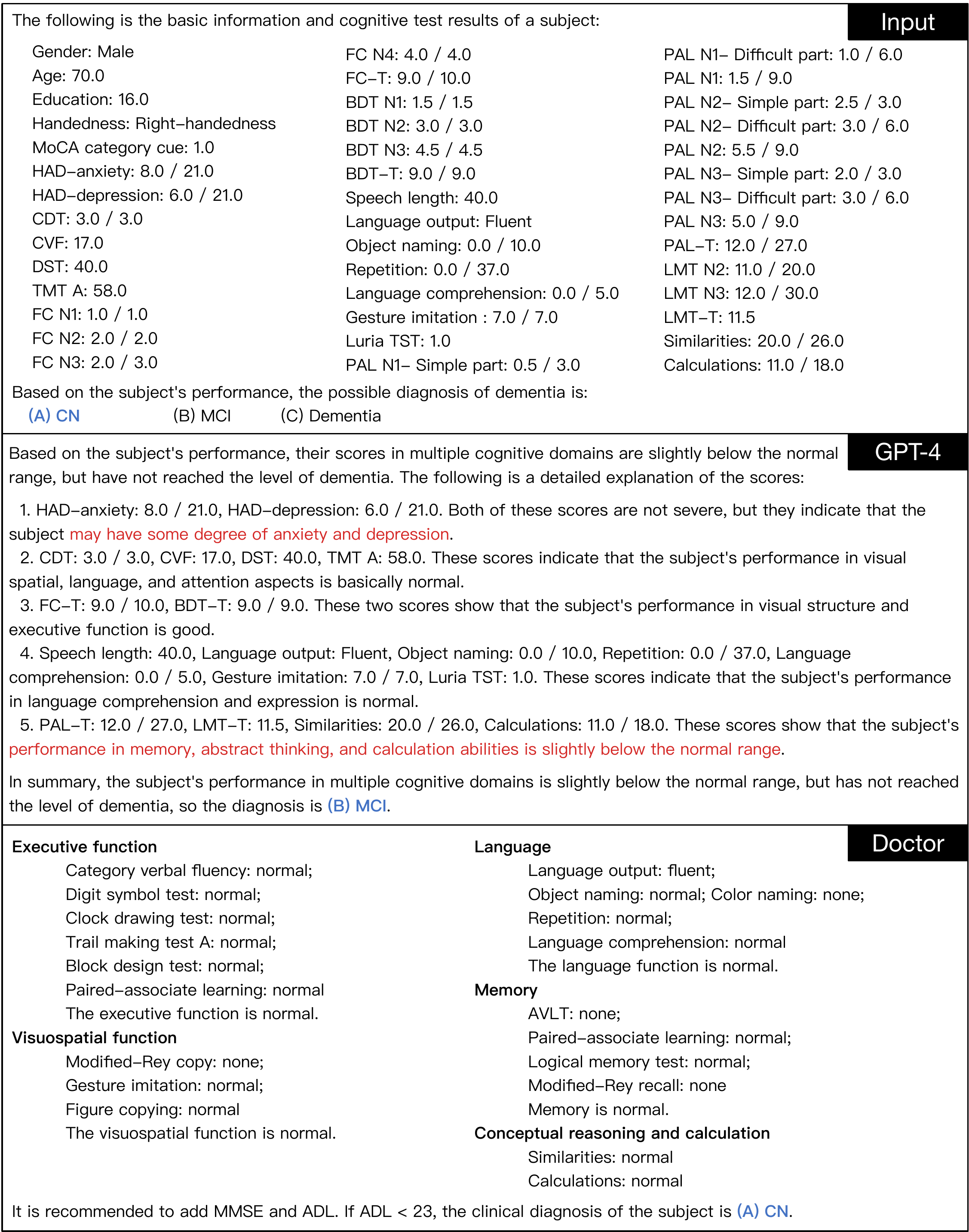}
    \caption{The second example of a comparison between GPT-4 and a doctor's diagnosis on the PUMCH dataset (English translation). The Chinese original text is shown in Figure \ref{fig:pumch_case_cn_1}, and a detailed description of each feature can be found in Table \ref{tab:feature-table}.}
    \label{fig:pumch_case_2}
\end{figure*}

The second example is shown in Figure \ref{fig:pumch_case_2}, with the format of each part being the same as that of Figure \ref{fig:pumch_case_1}. We can see that there is a disagreement between GPT-4 and the doctor in terms of diagnostic results for the subject in Figure \ref{fig:pumch_case_2}. GPT-4 misdiagnoses the CN subject as MCI, while the doctor correctly diagnoses the subject as CN. The reason for GPT-4's diagnosis is that the subject may have some degree of anxiety and depression, and his performance in memory, abstract thinking, and calculation abilities is slightly below the normal range. Although the subject has not reached the level of dementia, GPT-4 tends to diagnose him as MCI. On the other hand, the doctor believes that all the subject's test results are normal and suggests adding MMSE and ADL. If the ADL score is less than 23, the subject is considered completely normal. Comparing GPT-4 with doctors, we find that GPT-4 has different criteria for determining whether each test is abnormal. Doctors generally use the cut-off values corresponding to each test as the basis for judgment, while GPT-4 may not be able to fully match the test with its corresponding cut-off values, resulting in different judgments on individual tests compared to the doctor. In addition, GPT-4's preference for sequential analysis leads to a less accurate assessment of the subject's overall condition compared to the doctor. Finally, we can also find that doctors can expand their professional knowledge, such as seeking the ADL result that is not in the input, while GPT-4 is more limited to the existing input.

In the experiment, we also observe that GPT-4's diagnosis is greatly influenced by the input, which puts higher demands on the quality of the input. For example, if we input the wrong total score for a test, its judgment on that test will be severely affected, and such errors may not be directly detected and corrected from GPT-4's results. There are also some tests that have a maximum score, but a higher score is not necessarily better, and GPT-4 may mistakenly think the subject's cognitive function is impaired due to the low score.

\section{Discussion}
The present study finds that although some research claims that large language models like GPT-4 exhibit human-level performance on various professional and academic benchmarks\citep{bubeck2023sparks,openai2023gpt4,nori2023capabilities}, they still cannot outperform traditional AI tools in dementia diagnosis and prediction tasks. This finding contradicts some current research findings\citep{nori2023capabilities,kasai2023evaluating,he2023will}, mainly due to our use of private datasets and more challenging tasks in our study.

We conduct experiments on two real clinical datasets and use the private dataset PUMCH to avoid information leakage (leakage effects)\citep{inan2021training,saravanan2018data}, thereby more accurately measuring GPT-4's ability in dementia diagnosis and prediction. Although many related works have investigated the capabilities of LLMs like GPT-4 in specialized domains, most of them use public datasets \citep{kung2023performance,jin2021disease,jin2019pubmedqa,pal2022medmcqa}, making it difficult to avoid inflated results due to information leakage. For example, compared with other traditional AI tools, GPT-4's performance on the PUMCH-T dataset is far worse than its performance on the ADNI dataset. Moreover, such information leakage is generally not intentional but introduced during the process of collecting corpora, making it difficult to avoid.

Furthermore, since we select dementia diagnosis and prediction problems in real-world settings, the tasks involve a large number of test results from different cognitive domains and require handling numerous numerical features, making the tasks themselves more challenging and better able to test the model's capabilities. Additionally, the diagnostic and prediction tasks we select differ from typical settings. For example, the task of the PUMCH dataset is to diagnose subjects with MMSE scores greater than or equal to 26, i.e., early diagnosis of dementia in a population considered cognitively normal by MMSE, which is much more difficult than general dementia diagnosis tasks.

The advantages and disadvantages of GPT-4 are exposed during the dementia diagnosis and prediction tasks, we summarize them as follows:

\textbf{GPT-4 Advantages}. Despite GPT-4 not yet surpassing traditional AI tools, it has many promising advantages. Firstly, we find that GPT-4 performs much better than expected in our experiments and could already match or outperform supervised learning models like Logistic Regression and Decision Tree in some scenarios under zero-shot or few-shot settings. This indicates that GPT-4 may be able to replace traditional machine learning models in tasks with limited training data in the future. 
Secondly, GPT-4 can utilize existing medical expertise for diagnosis. For example, we only need to tell GPT-4 the name of a cognitive test, and it will know which cognitive function the test corresponds to. On the contrary, traditional models can hardly obtain useful information just from the feature names.
Another advantage of GPT-4 is its ability to provide explanations for its decisions, a capability that most black-box models lack. In our case study, we conducted a qualitative analysis of GPT-4's interpretability and faithfulness by comparing its diagnostic basis with that of professional doctors. We find that the explanations provided by GPT-4 are highly readable and easy to understand. Moreover, for some correctly diagnosed cases, its diagnostic basis is not significantly different from that of doctors.

\textbf{GPT-4 Disadvantages}. GPT-4 also has some notable drawbacks. The first issue is that GPT-4 currently cannot be fine-tuned, making it difficult to fully utilize existing data in complex tasks like early dementia diagnosis, resulting in poor performance. The second issue is that GPT-4 has high requirements for input quality. In addition to designing specific prompt templates, feature names must be described and constrained appropriately. Otherwise, GPT-4 may misinterpret the input content, significantly affecting its performance. The outputs of GPT-4 are also sensitive to the prompt template in some cases, making minor modifications to the template may result in quite different results. The third issue is that GPT-4's ability to handle tabular data is still insufficient. This limitation prevents us from using table formats to save input length, thereby limiting the number of few-shot examples. The fourth issue is that although GPT-4 lists many reasons in its explanations, we cannot determine how these reasons contribute to the final diagnostic conclusion. In practice, we find that GPT-4's reasons and conclusions might be inconsistent, indicating that faithfulness cannot be guaranteed.

\textbf{Future Directions}. Future research could focus on addressing GPT-4's limitations, such as enabling fine-tuning for complex tasks, improving input requirements, enhancing tabular data handling, and ensuring faithfulness in explanations. Additionally, exploring the integration of GPT-4 with traditional AI tools to leverage their respective strengths could be a promising direction. It is also essential to investigate GPT-4's performance in other medical domains and tasks to better understand its potential in healthcare applications.

\textbf{Limitations}. One limitation of our work is that we only used two datasets, which may not fully represent the diversity of dementia diagnosis and prediction tasks. Moreover, our study focused on GPT-4 and GPT-3.5, and the findings may not generalize to other large language models. Further research should consider using a wider population and more diverse datasets and comparing the performance of different LLMs in similar tasks.

\section{Conclusion}
Our study provides valuable insights into the capabilities of large language models, specifically GPT-4, in the context of dementia diagnosis and prediction. 
To test GPT-4 accurately and fairly, we first design simple and effective prompt templates according to the tasks.
Our experimental results on two real clinical datasets indicate that, although GPT-4 has shown remarkable performance in some professional benchmarks, it does not currently outperform traditional AI tools in dementia diagnosis and prediction tasks. 
We also evaluate the interpretability and faithfulness of GPT-4 by comparing it with professional doctors.
Based on all the experimental results, we summarize the advantages and disadvantages of GPT-4 and propose future research directions.

\section{Acknowledgements}
Data collection and sharing for this project was funded by the Alzheimer's Disease Neuroimaging Initiative (ADNI) (National Institutes of Health Grant U01 AG024904) and DOD ADNI (Department of Defense award number W81XWH-12-2-0012). 
ADNI is funded by the National Institute on Aging, the National Institute of Biomedical Imaging and Bioengineering, and through generous contributions from the following: AbbVie, Alzheimer's Association; Alzheimer's Drug Discovery Foundation; Araclon Biotech; BioClinica, Inc.; Biogen; Bristol-Myers Squibb Company; CereSpir, Inc.; Cogstate; Eisai Inc.; Elan Pharmaceuticals, Inc.; Eli Lilly and Company; EuroImmun; F. Hoffmann-La Roche Ltd and its affiliated company Genentech, Inc.; Fujirebio; GE Healthcare; IXICO Ltd.; 
Janssen Alzheimer Immunotherapy Research \& Development, LLC.; Johnson \& Johnson Pharmaceutical Research \& Development LLC.; Lumosity; Lundbeck; Merck \& Co., Inc.; Meso Scale Diagnostics, LLC.; NeuroRx Research; Neurotrack Technologies; Novartis Pharmaceuticals Corporation; Pfizer Inc.; 
Piramal Imaging; Servier; Takeda Pharmaceutical Company; and Transition Therapeutics. The Canadian Institutes of Health Research is providing funds to support ADNI clinical sites in Canada. Private sector contributions are facilitated by the Foundation for the National Institutes of Health (www.fnih.org). The grantee organization is the Northern California Institute for Research and Education, and the study is coordinated by the Alzheimer's Therapeutic Research Institute at the University of Southern California. ADNI data are disseminated by the Laboratory for Neuro Imaging at the University of Southern California.

This work was supported in part by National Key Research and Development Program of China under Grant No. 2020YFA0804503, 2020YFA0804501, National Natural Science Foundation of China under Grant No. 62272264, 61521002, and Beijing Academy of Artificial Intelligence (BAAI).

\section{Ethics Approval and Consent to Participate}
All subjects gave their informed consent for inclusion before they participated in the study. The study was conducted in accordance with the Declaration of Helsinki, and the protocol was approved by the Ethics Committee of PUMCH (No. JS1836).

\bibliographystyle{ACM-Reference-Format}
\bibliography{sample-base}






\appendix
\section{Feature Description}
Table \ref{tab:feature-table} lists all the features used in the PUMCH dataset, including their Chinese names, English names, and detailed descriptions.

\begin{CJK*}{UTF8}{gbsn}
\begin{table}
  \caption{All the features used in the PUMCH dataset.}
  \label{tab:feature-table}
  \centering
  \resizebox{0.99\linewidth}{!}{
  \begin{tabular}{p{0.3\linewidth}p{0.3\linewidth}p{0.55\linewidth}}
    \toprule
Chinese Name	&	English Name	&	Description\\
    \midrule
AVLT第1次回忆	&	AVLT N1	&	The first learning trial of Auditory Verbal Learning Test (AVLT)\\
AVLT第2次回忆	&	AVLT N2	&	The second learning trial of AVLT\\
AVLT第3次回忆	&	AVLT N3	&	The third learning trial AVLT\\
AVLT短延迟回忆	&	AVLT N4	&	The fourth short delayed free recall trial of AVLT\\
AVLT长延迟回忆	&	AVLT N5	&	The fifth long delayed free recall trial of AVLT\\
AVLT线索回忆	&	AVLT N6	&	The sixth delayed category cue recall trial of AVLT\\
AVLT再认击中	&	AVLT-RH	&	Recognitions hits of AVLT\\
AVLT再认虚报	&	AVLT-RF	&	Recognitions false of AVLT\\
AVLT短时记忆	&	AVLT-L	&	Total score of three learning trials of AVLT\\
AVLT总分	&	AVLT-T	&	Total score of AVLT-L, AVLT N4 and AVLT N5\\
医院焦虑抑郁量表（HAD)焦虑得分	&	HAD-anxiety	&	Anxiety score of Hospital Anxiety and Depression scale\\
医院焦虑抑郁量表（HAD)抑郁得分	&	HAD-depression	&	Depression score of Hospital Anxiety and Depression scale\\
画钟测验	&	CDT	&	Clock drawing test\\
词语流畅性测验	&	CVF	&	Category verbal fluency\\
数字符号测验	&	DST	&	Digit symbol test\\
接龙测验A	&	TMT A	&	Trail making test A\\
接龙测验B	&	TMT B	&	Trail making test B\\
临摹图一得分	&	FC N1	&	The first figure of Figure Copying (FC)\\
临摹图二得分	&	FC N2	&	The second figure of FC\\
临摹图三得分	&	FC N3	&	The third figure of FC\\
临摹图四得分	&	FC N4	&	The fourth figure of FC\\
临摹总分	&	FC-T	&	The total score of four figures of FC\\
积木图二得分	&	BDT N1	&	The first figure of Block Design Test (BDT)\\
积木图三得分	&	BDT N2	&	The second figure of BDT\\
积木图四得分	&	BDT N3	&	The third figure of BDT\\
积木测验得分	&	BDT-T	&	The total score of BDT\\
语言甄别测试语句长度	&	Speech length	&	Sentence length of spontaneous speech \\
语句时间秒数	&	Speech time	&	Time of spontaneous speech\\
语义错语	&	Semantic paraphasia	&	Semantic paraphasia\\
语音错语	&	Phonemic paraphasia	&	Phonemic paraphasia\\
找词困难	&	Word retrieval	&	Hesitation and delay in spoken production\\
重复	&	Repetitive language	&	Repetitive language\\
语言流利程度	&	Language output	&	Language output\\
实物命名错误数	&	Object naming	&	The number of correctly named objects\\
颜色命名错误数	&	Color naming	&	The number of correctly named colors\\
总复述错误字数	&	Repetition	&	Repeating three sentences\\
总听指令执行错误数	&	Language comprehension	&	Executing five commands\\
单个动作模仿正确数	&	Gesture imitation 	&	Imitation of seven hand gestures\\
系列动作模仿正确数	&	Luria TST	&	Luria three-step task\\
联想学习第一次总分	&	PAL N1	&	The first learning trial of Paired-associate learning of The Clinical Memory Test (PAL)\\
联想学习第一次容易	&	PAL N1- Simple part	&	Six simple word pairs of PAL N1\\
联想学习第一次困难	&	PAL N1- Difficult part	&	Six difficult word pairs of PAL N1\\
联想学习第二次容易	&	PAL N2- Simple part	&	Six simple word pairs of PAL N2\\
联想学习第二次困难	&	PAL N2- Difficult part	&	Six difficult word pairs of PAL N2\\
联想学习第二次总分	&	PAL N2	&	The second learning trial of PAL\\
联想学习第三次容易	&	PAL N3- Simple part	&	Six simple word pairs of PAL N3\\
联想学习第三次困难	&	PAL N3- Difficult part	&	Six difficult word pairs of PAL N3\\
联想学习第三次总分	&	PAL N3	&	The third learning trial of PAL\\
联想学习三次总分	&	PAL-T	&	The total score of the three learning trials of PAL\\
情景记忆甲得分	&	LMT N1	&	The first story of logical memory test of modified Wechsler Memory Scale (LMT)\\
情景记忆乙得分	&	LMT N2	&	The second story of LMT\\
情景记忆丙得分	&	LMT N3	&	The third story of LMT\\
情景记忆总得分	&	LMT-T	&	The total score of LMT\\
相似性测验总分	&	Similarities	&	Similarities of the Wechsler Adult Intelligence Scale\\
计算总分	&	Calculations	&	Calculations of the Wechsler Adult Intelligence Scale\\
简化版Rey-Osterrieth复杂图形（Benson图形）-临摹分数	&	Modified-Rey copy	&	Copy of a modified Rey-Osterrieth figure\\
简化版Rey-Osterrieth复杂图形（Benson图形）-回忆分数	&	Modified-Rey recall	&	Modified Rey-Osterrieth figure with a 10-minute free recall\\
Modified Rey复杂图形再认(1正确；2错误）	&	Modified-Rey Recognition	&	Recognition of Modified Rey-Osterrieth figure\\
    \bottomrule
  \end{tabular}
  }
\end{table}
\end{CJK*}

\section{Case study}
\begin{figure*}
    \centering
    \includegraphics[width=\textwidth]{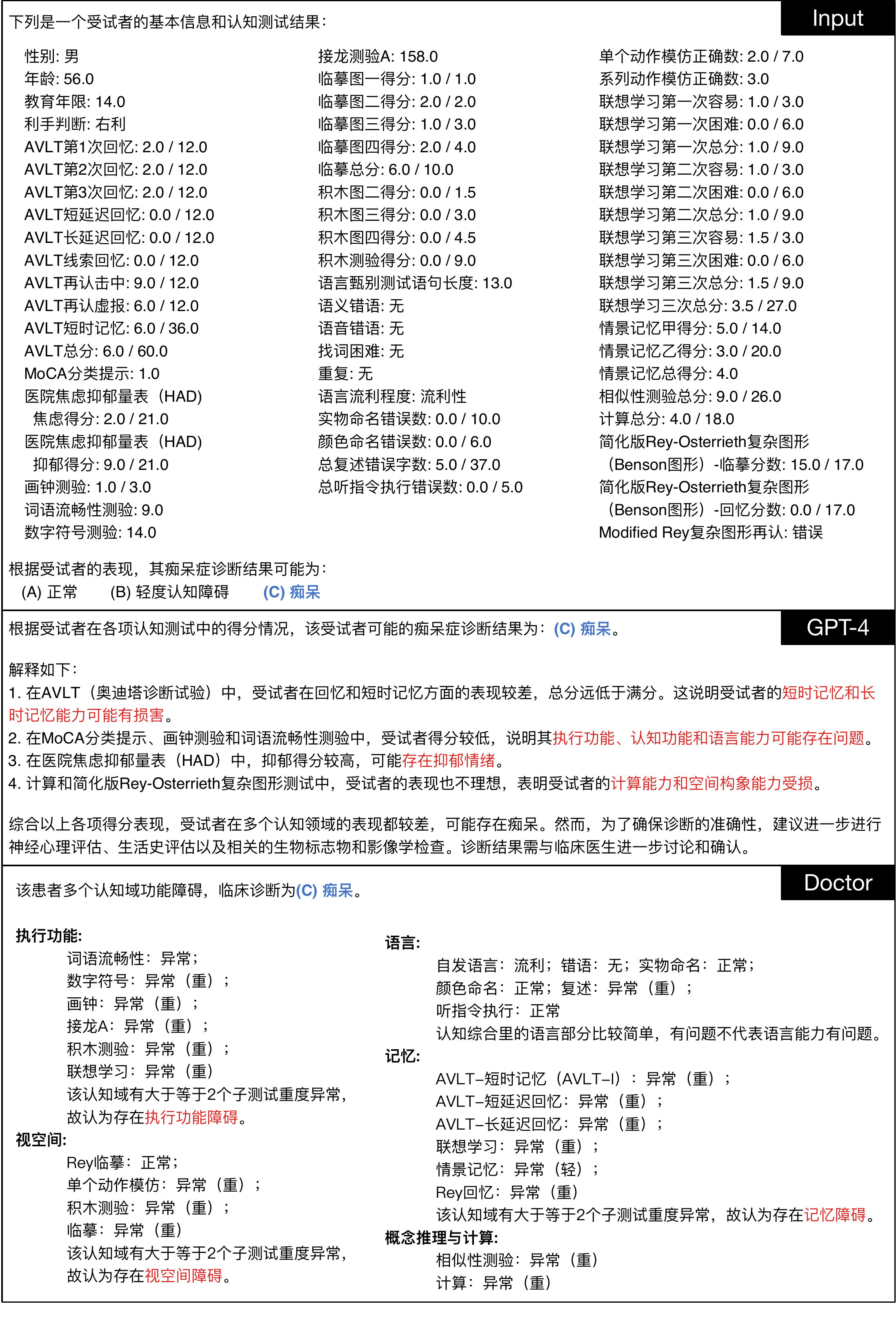}
    \caption{The first example of a comparison between GPT-4 and a doctor's diagnosis on the PUMCH dataset (Chinese original text). The detailed description of each feature can be found in Table \ref{tab:feature-table}.}
    \label{fig:pumch_case_cn_1}
\end{figure*}

\begin{figure*}
    \centering
    \includegraphics[width=\textwidth]{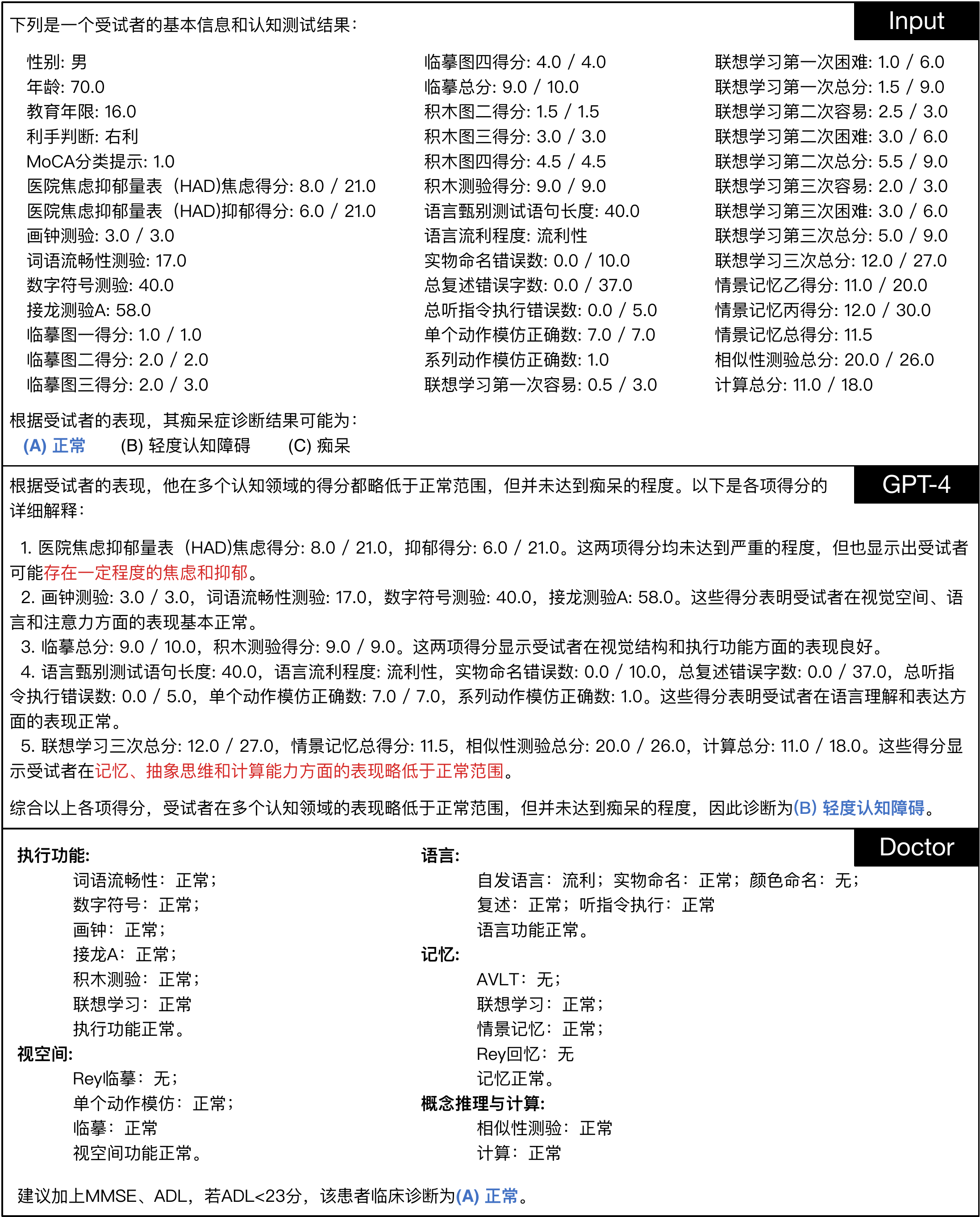}
    \caption{The second example of a comparison between GPT-4 and a doctor's diagnosis on the PUMCH dataset (Chinese original text). The detailed description of each feature can be found in Table \ref{tab:feature-table}.}
    \label{fig:pumch_case_cn_2}
\end{figure*}

Figures \ref{fig:pumch_case_cn_1} and \ref{fig:pumch_case_cn_2} show the first and second examples of a comparison between GPT-4 and a doctor's diagnosis on the PUMCH dataset (Chinese original text), respectively.
\end{document}